\documentclass{article}

\usepackage{epstopdf}
\usepackage{subfigure}

\usepackage[longnamesfirst,sort]{natbib}
\bibpunct[, ]{(}{)}{;}{a}{,}{,}
\renewcommand\bibfont{\fontsize{10}{12}\selectfont}

\usepackage{graphicx}

\usepackage[noend]{algorithmic}
\usepackage[ruled,vlined,linesnumbered]{algorithm2e}

\input epsf

\newcommand{\m}[1]{\texttt{#1}}
\newcommand{\hide}[1]{}

\newtheorem{example}{\vspace{-4mm}Example}

\begin{document}

\title{Exploiting the Pruning Power of Strong Local Consistencies Through Parallelization}

\author{Minas Dasygenis and Kostas Stergiou \\
Department of Informatics and Telecommunications Engineering \\
University of Western Macedonia \\ 
Kozani, Greece}

\maketitle

\begin{abstract}
Local consistencies stronger than arc consistency have received a lot of attention since the early days of CSP research. 
However, they have not been widely adopted by CSP solvers. This is because applying such consistencies can sometimes result in considerably smaller search tree sizes and therefore in important speed-ups, but in other cases the search space reduction may be small, causing severe run time penalties. Taking advantage of recent advances in parallelization, we propose a novel approach for the application of strong local consistencies (SLCs) that can improve their performance by largely preserving the speed-ups they offer in cases where they are successful, and eliminating the run time penalties in cases where they are unsuccessful. This approach is presented in the form of two search algorithms.
Both algorithms consist of a master search process, which is a typical CSP solver, and a number of slave processes, with each one implementing a SLC method. The first algorithm runs the different SLCs synchronously at each node of the search tree explored in the master process, while the second one can run them asynchronously at different nodes of the search tree. 
Experimental results demonstrate the benefits of the proposed method. 
\end{abstract}


\section{Introduction}

Constraint propagation is at the core of Constraint Programming (CP) and constitutes one the main reasons for its success. Constraint propagation algorithms typically enforce some local consistency property such as (generalized) arc consistency (G)AC or bounds consistency (BC) on the constraints of the problem, and in this way prune inconsistent values from the domains of the variables. Hence, algorithms for GAC and BC have been widely studied and applied. 

Local consistencies stronger than (G)AC have also received attention since they can offer even stronger pruning. Studies on such consistencies cover both binary constraints (e.g. the works by \cite{berlandier95,freuder96,db97,db01}) and non-binary constraints (e.g. the works by \cite{Janssen89,jegou93,vbd95a,BessiereStergiou08,relcon2010,lcv111}). Despite the wealth of research on strong local consistencies, and the recent advances in algorithms for such consistencies made among others by \cite{sac2011,cons2011,nic2012}, 
they have not been widely adopted by CSP solvers. This is because applying 
such consistencies can sometimes result in considerably smaller search tree sizes and therefore in important speed-ups, but in other cases the search space reduction may be small, causing severe run time penalties.

One inherent shortcoming of all such methods is that they have been designed for a sequential processing mode and therefore cannot exploit the very important recent advances in multicore parallel computing. These advances have triggered increasing interest in parallel constraint solving methods. But since local consistency algorithms are by nature sequential, meaning that their parallelization is quite challenging and requires very careful synchronization, there is limited, mainly theoretical, work on parallelizing such algorithms. 

In this paper we explore new ways to exploit the filtering offered by strong local consistencies through parallelization while keeping the complexity of synchronization manageable. Our goal is to exploit the extra filtering offered by strong local consistencies without penalizing the run time in cases where they are unsuccessful. Instead of trying to parallelize local consistency algorithms, we propose novel ways to apply different (strong) local consistencies during search in parallel to the main solver. 
The proposed approaches are presented in the form of search algorithms that consist of a master search process, which is a typical CSP solver, and a number of slave processes, which can implement strong local consistency algorithms. 

In the first algorithm, after each branching decision is made, the master process applies standard propagation (e.g. AC), while at the same time the slave processes are initiated. Each one of them applies some strong local consistency algorithm on a copy of the problem and is charged with delivering any value deletions that this algorithm makes back to the master process. All slave processes are terminated once propagation in the master process terminates, or alternatively, if a failure is detected by some slave process before propagation in the master process stops. In this way, even if a strong local consistency algorithm is not allowed to reach a fixpoint, it may still make many value deletions (or even discover failures) that the main propagation method, being weaker, cannot make. This method 
guarantees that the resulting solver will never be noticeably slower than a standard solver, 
even if the strong local consistencies employed do not offer any extra pruning.

The second method we propose 
again uses slave processes to run strong local consisteny algorithms. 
However, this method can run these algorithms asynchronously at different nodes of the search tree. The intuition behind this algorithm is the following: If the subtree rooted at some node of the search tree does not contain a solution then this can be verified either by 
searching the subtree or by applying a strong inference (e.g. local consistency) method at that node.  
We propose to apply both of them at the same time on different processes. 
Once a node is visited, the master process is responsible for running propagation and searching the subtree of the node if propagation does not fail, while the slave processes run strong local consistency algorithms at that node. If some slave process detects a failure then the master process is notified and a backtrack, which may be a large non-chronological one, occurs. 

Experiments with benchmark binary problems that were performed as a case study demonstrate that our algorithms outperform a standard CSP solver, sometimes by large margins, even when only one strong local consistency, i.e. maxRPC, is employed in addition to the standard AC propagation. Also, our algorithms outperform in quite a few cases a portfolio of two solvers running in parallel, where the first solver applies AC and the second maxRPC throughout search, but they are also outperformed in other cases. As a result,  by adding our algorithms to this portfolio, we can build a method that outperforms the portfolio of AC and maxRPC since on any given problem it is at least as good as any single solver or the portfolio of two solvers.

This paper is structured as follows. Section~\ref{sec:background} gives the necessary background. Sections~\ref{sec:sync} and~\ref{sec:async} present the two proposed algorithms and discuss 
possible extensions. In Section~\ref{sec:experiments} we give experimental results, while in Section~\ref{sec:related} we discuss related work. Finally, in Section~\ref{sec:conclusions} we conclude.

\section{Background}

\label{sec:background}

A {\em Constraint Satisfaction Problem} (CSP) is defined as a
triple $(X,D,C)$ where: 

\begin{itemize}

\item
$X=\{x_1,\ldots,x_n\}$ is a set of $n$ variables.

\item
$D=\{D(x_1),\ldots,D(x_n)\}$ is a set of ordered finite domains. Each domain $D(x_i)$, with $1\leq i \leq n$, contains the possible values for variable $x_i$.

\item
$C=\{c_1,\ldots,c_e\}$ is a set of $e$ constraints. Each constraint $c_i$ is a pair $(var(c_i),rel(c_i))$, where $var(c_i)=(x_{i_1},\ldots,x_{i_k})$ is an ordered subset of $X$, and $rel(c_i)$ contains the allowed combinations of values for the variables in $var(c_i)$. For simplicity, a binary constraint between two variables $x_i$ and $x_j$ will be denoted by $c_{ij}$.

\end{itemize}

The concept of {\em local consistency} (LC) is central to CP. LCs are used prior to and during search through what is known as constraint propagation to filter domains and discover inconsistencies early. The most widely studied LC is arc consistency. A binary CSP is {\em Arc Consistent} (AC) iff for all $x_i \in X$, $D(x_i)$ is non-empty and all values of $D(x_i)$ are AC. A value $a\in D(x_i)$ is AC iff for any constraint $c_{ij} \in C$, there exists at least one value $b \in D(x_j)$ s.t. the assignments $x_i=a$ and $x_j=b$ are consistent (i.e. they satisfy $c_{ij}$). In this case $b$ is called a {\em support} for $a$. The generalization of AC to non-binary constraints is known as GAC. 

Apart from AC and GAC, numerous other LCs have been proposed for binary and non-binary constraints. Here we are particularly interested in LCs that are stronger, i.e. can achieve stronger pruning, than AC. In the following we will simply refer to such LCs as {\em strong LCs}. 

Since the experiments included in this paper concern binary problems, we will focus on LCs for binary constraints hereafter. However, the algorithms presented below are generic and do not depend on the arity of the constraints.

LCs that only prune values from domains and do not add new constraints or alter existing ones are called {\em domain filtering} consistencies by \cite{db01}. One of the most efficient such LCs that was proposed by \cite{db97} is maxRPC. 
A binary CSP is {\em max Restricted Path Consistent} (maxRPC) iff it is AC and for each value $a \in D(x_i)$ and variable $x_j$ constrained with $x_i$, there is a support $b$ for $a$ in $D(x_j)$ s.t. the pair of values $(a,b)$ is path consistent. That is, for any third variable $x_k$ constrained with $x_i$ and $x_j$ there exists a value $c \in D(x_k)$ that is consistent with both $a$ and $b$. 

{\em Singleton Arc Consistent} (SAC) is another strong LC that was proposed by \cite{db01}. A binary CSP is SAC iff it has non-empty domains and for any assignment $x_i=a$ of a variable $x_i \in X$, the resulting subproblem, denoted by $P_{x_i=a}$, can be made AC. If $P_{x_i=a}$ cannot be made AC, SAC removes $a$ from $D(x_i)$.

Backtracking tree search is the standard complete method for solving CSPs. This method interleaves branching decisions (e.g. variable assignments) 
with constraint propagation. A backtracking algorithm searches for a solution in the space of possible variable assignments by gradually extending a partial assignment until it becomes a solution or proves that no solution exists. 

The description of the algorithms in the next section follows a classical \emph{d}-way branching scheme where the values of any variable are tried one by one\footnote{This is not a requirement of the algorithms since they are equally applicable with any other branching scheme.}. Given \emph{d}-way branching, the root of the search tree that the algorithm builds corresponds to the initial empty assignment (typically preprocessing is applied at the root) 
and thereafter, level \emph{i} of the tree corresponds to the \emph{i}$^{th}$ selected variable. Each node of the tree at level \emph{i} corresponds to a partial assignment starting from the first selected variable down to the \emph{i}$^{th}$ variable. 

After each variable assignment, a constraint propagation phase follows. Typically, this consists of applying LC algorithms on the constraints of the problem (e.g. AC on binary constraints). CP solvers usually implement constraint propagation using a queue where entities of the problem (i.e. variables, or constraints, or propagators) are inserted once an event such as a value removal occurs. Then the elements of the queue are iteratively removed and domains are revised (i.e. values that are no longer consistent are removed from domains). This may cause new queue insertions and so on, until the queue becomes empty or a failure occurs. The latter takes place when propagation removes all values from the domain of a variable. This is known as a {\em domain wipeout} (DWO). If a DWO is detected then the algorithm rejects the latest assignment and tries the next available one. The search algorithm which applies AC after the branching decisions is known as {\em Maintaining Arc Consistency} (MAC).

Algorithms such as MAC employ {\em chronological} backtracking. That is, if all possible assignments of the currently selected variable at search tree level \emph{i} have been tried and failed, in which case we have a {\em dead-end}, then the algorithm moves back to the previously selected variable at level \emph{i-1} and tries its next value. A {\em non-chronological} bakcktrack, or {\em backjump}, occurs when after a dead-end is encountered the search algorithm moves further up the search tree, instead of moving to level \emph{i-1}. A number of algorithms that allow for backjumps were proposed in the past (e.g. CBJ by \cite{prosser93}), and although backjumping had been neglected for quite a few years, it has recently gathered attention again through ideas such as lazy clause learning by \cite{Ohrimenko09}. As we will explain, one of the methods we propose allows for backjumps in a novel way by exploiting the pruning power of strong LCs and parallelization.

\section{A Synchronous Algorithm}

\label{sec:sync}

We now describe our first search algorithm that incorporates the parallel application of different LCs. We assume that there are \m{k} LCs available (\m{LC$_1$,LC$_2$,...,LC$_k$}) and that LC$_1$, i.e. the default propagation method of the solver, is AC. The search process of the solver, i.e. variable assignments, domain updates, backtracks, and AC propagation, runs on a single thread (the {\em main} or {\em master} thread). This means that the search algorithm explores a single search tree. As we will explain, strong LCs are used 
to cut down the size of the search tree and hence speed up the solving process.

Algorithm~\ref{synch} describes a synchronous method for applying strong LCs in parallel. This method aims at exploiting the filtering power offered by strong LCs, to some extent, without slowing down the solver in cases where there is little or no extra pruning. This is done by applying the main propagation method and the strong LCs {\em synchronously} at each node of the search tree. In the following we will call this algorithm {\em Sync}.


Specifically, Algorithm~\ref{synch} implements the following simple idea: In addition to the application of AC after each branching decision, the remaining \m{k-1} LCs are also applied by running them on different threads in parallel. As soon as AC reaches a fixpoint, or any of the LCs detects a DWO, all threads where the LCs are executed 
are stopped. Given that the LCs applied are stronger than AC, and therefore the algorithms that enforce them are more expensive than a typical AC algorithm, in most of their invocations they will not be allowed to reach a fixpoint since AC will reach a fixpoint earlier. Despite this, it is quite possible that during the available run time they will prune some values that AC cannot prune, and they may even discover DWOs that AC cannot detect. 

\begin{algorithm}[htbp]
\KwIn{a CSP instance \m{P}}
\KwOut{\m{SOLUTION FOUND} or \m{FAILURE}}
\Begin {
	run \m{LC$_1$,LC$_2$,...,LC$_k$} on different threads\\
   {\bf until} a DWO is detected {\bf or} LC$_1$ reaches fixpoint
}
\If {a DWO has been detected} {
	 \Return{\m{FAILURE}}
}
\Else { 
	merge the value deletions made by the \m{k} LCs
}
\m{tree$\_$level} $\leftarrow$ 1\\
select the first variable \m{x$_{\m{tree\_level}}$}\\
\While {1 $\leq$ \m{tree$\_$level}$\leq$ \m{n}}	{
	\While {there are values left in \m{D(x$_{\m{tree\_level}}$)}} {
		select the next value \m{a} for \m{x$_{\m{tree\_level}}$}\\
		remove all other values from \m{D(x$_{\m{tree\_level}}$)}\\
		\Begin {
			run \m{LC$_1$,LC$_2$,...,LC$_k$} on different threads\\
			{\bf until} a DWO is detected {\bf or} LC$_1$ reaches fixpoint
		}		 
	   \If {a DWO has been detected} {
			value \m{a} of \m{x$_{\m{tree\_level}}$} is rejected\\ 
			domains are restored
		}
		\Else { 
			merge the value deletions made by the \m{k} LCs\\
			{\bf break}
		}
	}
	\If {there is no value left for \m{x$_{\m{tree\_level}}$}} {	
		\m{tree$\_$level} $\leftarrow$ \m{tree$\_$level} - 1 // BACKTRACK
   }
	\Else {
		assign value \m{a} to \m{x$_{\m{tree\_level}}$}\\ 
		\m{tree$\_$level} $\leftarrow$ \m{tree$\_$level} + 1 // GO FORWARD\\
		select the next variable \m{x$_{\m{tree\_level}}$}	
	}
}
\If {\m{tree$\_$level} is 0} {
	\Return{\m{FAILURE}}
}
\Else {
	\Return{\m{SOLUTION FOUND}}
}
\caption{\m{Synchronous Algorithm}} \label{synch}
\end{algorithm}

Algorithm {\em Sync} is based on a standard iterative description of a MAC-like search algorithm. \m{tree$\_$level} denotes the current level of the search tree. The algorithm iteratively searches the space of possible variable assignments, through the while loop in line 10, until a solution is found or it is proved that none exists.

The parallel invocations of the different LCs occur in lines 1-3 (preprocessing) and 14-16 (after each branching decision). Hence, all the LCs are run in parallel at each node of the search tree, starting from the root. Note that any LC other than AC operates on a copy of the variables' domains. Importantly, if the threads running the LCs are stopped without having detected a DWO then all value deletions caused by the different LCs are merged (lines 7 and 21). That is, they are carried over from the copies to the domains of the main thread where AC is run.

{\em Sync} starts by preprocessing the given CSP instance (lines 1-7). This is done by running all the available LCs in parallel. 
If some LC detects a DWO then all threads are stopped since the problem has been proved to be insoluble (line 4). Otherwise, the threads are terminated once AC has been applied on the problem, and the value deletions made by the various LCs are carried over to the domains of the main thread (line 7). 

Thereafter, search commences by setting \m{tree$\_$level} is set to 1 (line 8) and asking the variable ordering heuristic to make its first choice of variable (line 9). The loop of line 10 iterates over the variables, while the loop of line 11 iterates over the values of the selected variable \m{x$_{\m{tree\_level}}$}. After making a value assignment (lines 12-13), the algorithm enters the constraint propagation phase. At this point the variables' domains are copied and the various LC algorithms are run on different threads on these copies (line 15). Constraint propagation stops when AC has been applied on the main thread or when a DWO is detected in any of the threads. In the first case the value deletions made by the various LC algorithms are copied to the main thread (line 21), while in the latter case the current value assignment is rejected and a new value is tried for the current variable \m{x$_{\m{tree\_level}}$} in the next iteration of the inner while loop.

If the propagation of all value assignments for the current variable fail then a chronological backtrack to the immediately preceding variable is triggered (lines 23-24). Otherwise, if a value assignment is successfully propagated then the algorithm moves forward by selecting one of the unassigned variables (lines 26-28).

Significantly, the synchronization process is trivial to implement and the overheads for copying domains are negligible. Hence, the algorithm will not be noticeably slower than a standard sequential solver even if the strong LCs do not offer any extra pruning. 

\section{An Asynchronous Algorithm}

\label{sec:async}

Algorithm~\ref{asynch} describes an asynchronous method for applying strong LCs, , where ``asynchronous'' refers to the simultaneous application of LC algorithms at different nodes of the tree. In the following we will call this algorithm {\em Async}. The main difference with Algorithm~\ref{synch} is that LC algorithms that run on different processes are allowed to reach their fixpoints and can run at different nodes of the search tree at the same time. This can utilize the filtering power of strong LCs to a greater extent. 

\begin{algorithm}[htbp]
\KwIn{a CSP instance \m{P}}
\KwOut{\m{SOLUTION FOUND} or \m{FAILURE}}

\For {$i=1...k$} {
	tree$\_$level$_i$ $\leftarrow$ 0\\
	\m{state}$_i$ $\leftarrow$ \m{RUNNING}\\ 
}
\Begin {
	run \m{LC$_1$,LC$_2$,...,LC$_k$} on different processes\\
}
\If {a DWO has been detected} {
	\Return{\m{FAILURE}}
}
\m{tree$\_$level$_1$} $\leftarrow$ 1\\
select the first variable \m{x$_{\m{tree\_level$_1$}}$}\\
\While {1 $\leq$ \m{tree$\_$level$_1$} $\leq$ \m{n}} {
	\While {there are values left in \m{D(x$_{\m{tree\_level$_1$}}$)}} {
		\m{fail} $\leftarrow$ FALSE\\
\m{pa} = \m{$\{$x$_1$=a$_1$,x$_2$=a$_2$,...,x$_s$=a$_s\}$} $\leftarrow$ the current partial assignment\\
\m{BT$\_$level} $\leftarrow$ \m{ForceBT(P,pa)}\\
\If {\m{BT$\_$level} $<$ \m{tree$\_$level$_1$}} {
	\m{fail} $\leftarrow$ TRUE\\
	{\bf break}
}
		select the next value \m{a} for \m{x$_{\m{tree\_level$_1$}}$} and remove all other values from 	\m{D(x$_{\m{tree\_level$_1$}}$)}\\
		\For {any \m{LC$_i$}(tree$\_$level$_i$) ($1\leq i \leq k$) s.t. \m{state}$_i$=\m{IDLE}} {
			\m{state}$_i$ $\leftarrow$ \m{RUNNING} \\
			run \m{LC$_i$(tree$\_$level$_1$)} on a different process
		}
		\If {a DWO has been detected by any LC called at tree$\_$level$_1$} 	   
 		{
			value \m{a} of \m{x$_{\m{tree\_level$_1$}}$} is rejected\\ 				
			\m{fail} $\leftarrow$ TRUE\\
			\If {there is no value left for \m{x$_{\m{tree\_level$_1$}}$}} 
			{
				\m{BT$\_$level} $\leftarrow$ \m{tree$\_$level$_1$} - 1
			}		
		}
		{\bf else break}
	}
	\If {\m{fail} = TRUE}	{
		\m{tree$\_$level$_1$} $\leftarrow$ \m{BT$\_$level} \\
		domains are restored\\
		\For {any \m{LC$_i$}(tree$\_$level$_i$) ($1\leq i \leq k$) s.t. \m{state}$_i$=\m{RUNNING} {\bf and} \m{tree$\_$level$_i$}$\geq$\m{tree$\_$level$_1$}}	{
			stop the process of \m{LC$_i$}\\
			\m{state}$_i$ $\leftarrow$ \m{IDLE} 
		}
   }	
	\Else {
		assign value \m{a} to \m{x$_{\m{tree\_level$_1$}}$}\\ 
		\m{tree$\_$level$_1$} $\leftarrow$ \m{tree$\_$level$_1$} + 1\\ 
		select the next variable \m{x$_{\m{tree\_level$_1$}}$}	
	}
}
{\bf if} \m{tree$\_$level$_1$} is 0 {\bf then return} \m{FAILURE} \\
{\bf else return} \m{SOLUTION FOUND}
\caption{\m{Asynchronous Algorithm}} \label{asynch}
\end{algorithm}

\begin{function}
\m{BT$\_$level1} $\leftarrow$ \m{BT$\_$level2} $\leftarrow$ \m{n+1}\\
$\{$\m{LC}$_g$(tree$\_$level$_g$),...,\m{LC}$_h$(tree$\_$level$_h$)$\}$ $\leftarrow$ the set of LC calls that have detected a DWO \\
\If {$\{$\m{LC}$_g$(tree$\_$level$_g$),...,\m{LC}$_h$(tree$\_$level$_h$)$\} \neq \emptyset$} {
	\m{BT$\_$level1} $\leftarrow$ \m{min}$\{$tree$\_$level$_g$,...,tree$\_$level$_h\}$
} 
$\{$\m{LC}$_{g'}$(tree$\_$level$_{g'}$),...,\m{LC}$_{h'}$(tree$\_$level$_{h'}$)$\}$ $\leftarrow$ the set of LC calls that have deleted a value \m{a$_i$} s.t. \m{x$_i$=a$_i$} $\in$ \m{pa} \\
\If {$\{$\m{LC}$_{g'}$(tree$\_$level$_{g'}$),...,\m{LC}$_{h'}$(tree$\_$level$_{h'}$)$\}$ $\neq \emptyset$} {
	\m{BT$\_$level2} $\leftarrow$ \m{min}(tree$\_$level$_{g'}$,...,tree$\_$level$_{h'}$)
}
\Return{\m{min}(\m{tree$\_$level$_1$},\m{BT$\_$level1},\m{BT$\_$level2})}

\caption{ForceBT(\m{P}:a CSP instance,\m{pa}:a partial assignment)} \label{func}
\end{function}

In each call to some LC$_i$ we pass as an argument \m{tree$\_$level$_i$}: the level of the search tree where the call is made. Since LC$_1$ is AC, \m{tree$\_$level$_1$} will always denote the current level of the search tree for the master process where the search mechanism of the solver operates. There is also a variable \m{state}$_i$ associated with each strong LC$_i$. This variable can take values  RUNNING and IDLE. If \m{state}$_i$=RUNNING then a process applying LC$_i$ is currently running. Otherwise, if \m{state}$_i$=IDLE then no process applying LC$_i$ is currently running. Once the application of some LC$_i$ terminates then \m{state}$_i$ is automatically set to IDLE.

Algorithm {\em Async} starts by applying all the LCs simultaneously at the root of the search tree (preprocessing at lines 1-5). As in {\em Sync}, all LCs apart from AC operate on copies of the variables' domains.  But in contrast to {\em Sync}, the threads running the strong LCs are not stopped once AC reaches a fixpoint but they continue their execution. 
Once AC finishes with preprocessing, the algorithm checks if any LC algorithm 
has detected a DWO (line 6). In such a case, the problem has been proved to be insoluble. If no DWO is detected, then search, which is running on the master process, kicks off (lines 9-10). In the meantime, some or all of the processes running strong LCs at the root of the tree will continue their execution until they reach a fixpoint. 

Before a new variable assignment is made, the algorithm checks if any strong LC algorithms have finished processing and if so, whether a backtrack can be initiated. 
This is done by calling Function \m{ForceBT} with the current partial assignment as an argument. There are two cases where a (non-chronological) backtrack can be forced:
\begin{enumerate}
\item Some LC algorithms that were called at levels shallower than the current level have detected a DWO. 
\item Some LC algorithms that were called at levels shallower than the current level have deleted a value that is part of the current partial assignment.
\end{enumerate}
If any of these cases, which we call {\em reasons for failure}, occurs, Function \m{ForceBT} compares the levels of the search tree where the relevant LC algorithms were called (lines 3-4 and 6-7) and the shallowest level (i.e. the one closest to the root) is returned in line 8. This will then trigger a {\em non-chronological backtrack} of the search mechanism running on the master process. If none of the two cases occurs then the current search level \m{tree$\_$level$_1$} is returned.

If \m{ForceBT} returns \m{tree$\_$level$_1$} then the selected variable assignment is temporarily made (line 18). Then AC propagation is run on the master process and at the same time, any LC that is at an IDLE state is set to RUNNING and is run in parallel to AC (lines 19-21). Once AC finishes, the algorithm checks if any of these LCs has also terminated.  For any LC that has terminated, including AC, the algorithm then checks if it has detected a DWO or not (line 22). If a DWO has been detected, the current variable assignment is rejected and if there are no values left for the current variable, \m{BT$\_$level} is set to the current level minus 1 to cause a chronological backtrack. Otherwise, the algorithm exits the inner while loop of line 11 and proceeds to make the currently tried assignment, move to the next level of the search tree, and select the next variable (lines 35-38).  

The way reasons for failure are handled after DWOs or deletions from the partial assignment occur (lines 29-34) is where we gain from the application of the strong LCs. There are three cases:

\begin{enumerate}

\item None of the strong LCs that finished running within the current iteration detected a reason for failure at a level shallower than the current one. This means that the failure was detected by AC or another LC in line 22. If there are still values left for the current variable \m{x}$_{\m{tree\_level$_1$}}$ then \m{BT$\_$level} will be set to  \m{tree$\_$level$_1$} (from the call to \m{ForceBT}) and therefore the algorithm will not backtrack and will proceed to try the next value of \m{x}$_{\m{tree\_level$_1$}}$ in the next iteration.


\item The failure was detected by AC or another LC in line 22 but no values are left for the current variable. In this case a chronological backtrack will take place since \m{BT$\_$level} will be set to \m{tree$\_$level$_1$} - 1 (from line 26).


\item Some of the strong LCs that finished running within the current iteration  detected a reason for failure at a level shallower than the current one. In this case a non-chronological backtrack (i.e. a backjump) may take place since \m{ForceBT} will set \m{BT$\_$level} to the shallowest level where a reason for failure was detected. 

\end{enumerate}

After the backtrack point \m{BT$\_$level} has been determined, a crucial step follows. Namely, all slave processes running a strong LC at a level equal or greater than \m{BT$\_$level} are stopped and their state is set to IDLE (lines 32-34). This is because the subtree below the node where they were called has been proved not to contain a solution (hence the backtrack). Therefore, continuing their execution is fruitless.

The following example illustrates the algorithm.

\begin{figure}
\begin{center}
\includegraphics[width=0.75\textwidth]{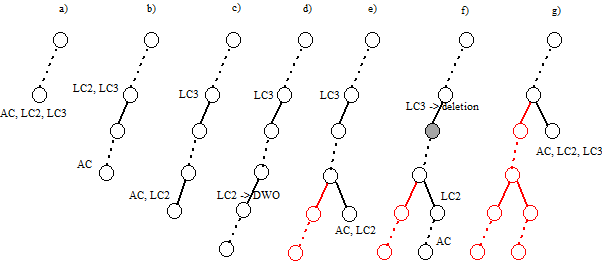} 
\caption{
A solid (resp. dashed) line between two nodes denotes a parent-child (resp. ancestor-descendant) relationship. At any point the shaded part of the search tree has been already explored.} \label{tree}
\end{center}
\end{figure}


\begin{example}\rm
Assume that two strong LCs, LC2 and LC3, are utilized in addition to AC by Algorithm \ref{asynch}. 
In Figure \ref{tree}a) the algorithm initiates the master and slave processes for the application of the three LCs at some node of the search tree. In Figure \ref{tree}b) LC2 and LC3 are still running but search on the master process has moved on and AC is applied at a node further down the search tree. In Figure \ref{tree}c) LC2 has terminated without disovering a reason for failure before the next assignment is made. Hence, after the next assigment is made, AC and LC2 are applied simultaneously. In Figure \ref{tree}d) search has moved further down the search tree but before applying AC at the current node, LC2 terminates by detecting a DWO. At this point a non-chronological backtrack is initiated, and as Figure \ref{tree}e) illustrates, AC and LC2 are applied in parallel at the sibling of the node where the DWO was detected. In Figure \ref{tree}f) LC2 is still being applied at that node while search has moved on and AC is applied at a node further down. But at this time LC3 terminates having deleted a value that belongs to the partial assignment (the shaded node).
Hence, in Figure \ref{tree}g) a non-chronological backtrack up to the level where this assignment was made will occur. Then all three LCs will be again applied.  
\end{example}

\subsection{Extensions of the algorithms}

One shortcoming of the algorithms, as they are describe above, is that only a portion of the parallelization power offered by modern machines is utilized by them. This is because the algorithms dedicate one core to each different LC, and realistically there will only be a few LCs available in any CSP solver. Even in the case of binary constraints, where quite a few different LCs have been proposed, typical solvers only include a basic propagation mechanism (e.g. AC) and perhaps one or two stronger methods such as SAC and maxRPC. 

We now briefly discuss two ideas, one for each algorithm, that can be exploited to overcome this limitation.  We intend to implement and test them in the future. 

\subsubsection{Sync}

Regarding algorithm {\em Sync}, the full potential of multithreading can be exploited using different orderings of the propagation queue. Such orderings can be generated  through randomization as we will explain. First of all, it is well known that the order in which elements of the propagation queue (e.g. variables or constraints) are processed has an impact on the cpu time that propagation takes. There have been quite a few studies on this and a number of heuristics for efficiently ordering the elements of the queue, mainly in the case of arc consistency, have been proposed by \cite{wallace92,boussem04,balsterg08}. Of course, any different ordering of the queue results in exactly the same value deletions if the particular propagation algorithm is run to completion (i.e. until its fixpoint is reached). However, if the algorithm is not run to completion, as is the case for strong LC algorithms in the framework of {\em Sync}, different queue orderings may result in different value deletions. 

For example, assume that an algorithm that applies maxRPC removes a variable $x_i$ from the queue. Then all values of all variables constrained with $x_i$ must be checked for maxRPC. In our framework, if this algorithm is run in parallel to the main AC propagation mechanism, it will be stopped once AC terminates. Therefore, it is likely that only some of the variables constrained with $x_i$ may have been processed until the algorithm is stopped. If these variables are handled in parallel threads, in different orderings, then potentially different value deletions may be made in each thread. 

To summarize, each time the application of a strong LC is initiated on some thread within algorithm {\em Sync}, we can additionally allocate any number of available threads where the same LC is run under different (randomized) queue orderings. Once constraint propagation on the main thread terminates, all the threads are stopped and any value deletions made are merged. 

\subsubsection{Async}

Assuming that {\em k-1} strong LC are available in addition to the standard propagation mechanism, algorithm {\em Async} will utilize at most {\em k} threads as at any time during search. 
One way to utilize all of the available threads is the following: At lines 19-21 of Algorithm \ref{asynch}, instead of initiating the application only of those LCs whose state is IDLE, we can initiate the application of all LCs as long as there are available resources. This will mean that at any time during search the same LC may run simultaneously at many different nodes of the search tree. 

For example, consider a case where there is only one strong LC available (say maxRPC). {\em Async} will start by applying AC and maxRPC at the root of the search tree. Once the AC algorithm terminates, and assuming no DWO occurs, search on the main thread will start by assigning a variable $x$ and again applying AC, while at the same time the maxRPC algorithm may still be running at the root node. The extension of {\em Async} discussed here will initiate the application of maxRPC on one of the avaiable threads, in addition to AC, after the assignment of $x$. Hence, the maxRPC algorithm may now be running at two different nodes at the same time. 

This extension of {\em Async} should result in faster search, but on the other hand a greater effort for synchronization will be required.

\section{Experimental Results}

\label{sec:experiments}


As a case study, we have experimented with benchmark binary CSPs taken from Christophe Lecoutre's XCSP repository and used in CSP solvers competitions. Specifically, we experimented with 300 instances belonging to the following classes: radio links frequency assignment (rlfap), graph coloring (gc), driver (dr), forced random (frb), quasigroup completion (qcp), quasigroups with holes (qwh).

All experiments were performed in a multiprocessor shared memory
system consisting of 8 cores (the total number with hyperthreading is 16)
of Intel(R) Xeon(R) CPU E5520 at 2.27GHz
and with 78GB of memory, under the operating system 
Ubuntu Linux 12.04.
The framework that we used for the implementation was the OpenMP 3.0 \cite{openmp08}
API extensions on the GCC 4.6.3 compiler , without any compiler optimization parameters. 
To assess the performance we used the practical execution time (or wall clock) as a measure.
The practical execution time (or wall clock) is the total time that a process requires in order to complete its computation. The execution time is obtained by calling the C POSIX.1-2001 function {\tt clock$\_$gettime()} and it is measured in nano seconds, with the highest available timer accuracy.

We used a standard MAC algorithm as the baseline solver, and in addition we implemented and applied 
maxRPC within the context of the proposed framework. Specifically, the two algorithms presented in Section 3 (denoted {\em Sync} and {\em Async} hereafter) apply 
maxRPC in slave processes parallel to the master process which applies AC. 
We compare {\em Sync} and {\em Async} to: 1) the baseline solver, i.e. MAC (denoted AC), and 2) a simple portfolio of two search algorithms that are run in parallel independently from one another. These algorithms respectively apply AC and maxRPC throughout search. For any given problem, the portfolio (denoted {\em pf$_{AC+maxRPC}$}) terminates once one of the algorithms finds a solution (or proves that none exists). Hence, for any given instance the portfolio gives the same result as either AC or maxRPC, depending on which of the two is better for the specific instance. Finally, we integrated {\em Sync} and {\em Async} within the portfolio resulting in a new portfolio that consists of four algorithms (denoted {\em pf$_{all}$}).

All algorithms used the dom/wdeg heuristic for variable ordering and lexicographical value ordering. Our algorithms were run 50 times on each instance and the median cpu times and node visits are reported\footnote{The mean values are quite close in general but are heavily influenced by a few outliers in some cases of soluble instances.}. A time limit of 1 hour was set.


\begin{table}[hbt]
\centering
\caption{Nodes (n) and cpu times (t) in seconds. The s and g prefixes stand for scen and graph respectively. 
The best cpu time for each instance is highlighted with bold.
} 
\begin{center}
\begin{footnotesize}
\begin{tabular}[hbt]{@{~}l@{~}@{~}c@{~}@{~}c@{~}@{~}c@{~}@{~}c@{~}@{~}c@{~}@{~}c@{~}@{~}c@{~}@{~}c@{~}@{~}c@{~}@{~}c@{~}@{~}}

\hline
instance & AC & & pf$_{AC+maxRPC}$ & & Async & & Sync & & pf$_{all}$ & \\

 & (n) & (t) & (n) & (t) & (n) & (t) & (n) & (t) & (n) & (t) \\
\hline

rlfap & & & & & & \\



s11-f12 & 7349 & 16 & 7349 & 17 & 5521 & 14 &  3983 & {\bf 10} & 3983 & {\bf 10} \\

\hline



s11-f10 & 9601 & 22 & 9601 & 24 & 7518 & 19 & 6198 & {\bf 15} & 6198 & {\bf 15} \\

\hline

s11-f09 & 101K & 295 & 101K & 299 & 73K & {\bf 246} & 80K & 277 & 73K & 248 \\

\hline

s02-f25 & 12688 & 11 & 12688 & 11 & 9024 & 8 & 5056 & {\bf 6} & 5056 & {\bf 6} \\

\hline



s03-f11 & 9486 & 14 & 9486 & 14 & 7082 & 12 & 5293 & {\bf 10} & 5293 & {\bf 10} \\

\hline

g08-f10 & 19590 & 33 & 8808 & 28 & 9715 & 25 & 7199 & {\bf 16} & 7199 & {\bf 16} \\

\hline

g14-f27 & 13833 & 10 & 4326 & 6 & 4869 & 6 & 4697 & {\bf 5} & 4697 & {\bf 5} \\



\hline

graph coloring & & & & & & \\

anna-8 & 69K & 18 & 69K & 18 & 46K & 16 & 30K & {\bf 10} & 30K & {\bf 10} \\

\hline





homer-8 & 69K & 99 & 69K & 101 & 50K & 80 & 29K & {\bf 44}  & 29K & 45 \\

\hline

ga120-7 & 65K & 7 & 25K & {\bf 4} & 45K & 9 & 29K & {\bf 4} & 29K & {\bf 4} \\

\hline

ga120-8 & 3208K & 310 & 1352K & 250 & 887K & {\bf 121} & 1999K
 & 211 & 887K & 122 \\

\hline

lei-450-8 & 107K & 784 & 107K & 787 & 80K & {\bf 560} & 86K & 630 & 80K & 562 \\

\hline

driver,frb & & & & & & \\

driver-8 & 3872 & 7 & 3872 & 7 & 2919 & {\bf 5} & 2742 & {\bf 5} & 2742 & {\bf 5} \\

\hline

driver-9 & 14129 & 66 & 14129 & 67 & 6534 & {\bf 31} & 11546 & 51 & 6534 & {\bf 31} \\

\hline

frb-35 & 26K & {\bf 8} & 26K & {\bf 8} & 20K & 9 & 24K & {\bf 8} & 24K & {\bf 8} \\

\hline

frb-40 & 45K & {\bf 16} & 45K & {\bf 16} & 32K & 18 & 39K & 17 & 45K & {\bf 16} \\

\hline

frb-45 & 1207K & 531 & 1207K & 532 & 601K & {\bf 310} & 1001K & 517  & 601K & 313 \\

\hline

qcp,qwh & & & & & & \\

qcp-15-0 & 102K & 71 & 21K & {\bf 29} & 58K & 46 & 85K & 63 & 21K & {\bf 29} \\

\hline

qcp-15-1 & 20988 & 17 & 3025 & 5 & 5325 & {\bf 4} & 8893 & 7 & 5325 & {\bf 4} \\

\hline





qcp-15-5 & 536K & 457 & 37K & {\bf 75} & 131K & 116 & 222K & 195 & 37K & 76 \\

\hline

qcp-15-6 & 62K & 47 & 62K & 47 & 29K & {\bf 24} & 49K & 38 & 29K & {\bf 24} \\

\hline



qcp-15-8 & 22K & 18 & 22K & 19 & 13K & {\bf 12} & 17K & 15 & 13K & {\bf 12} \\

\hline





qcp-15-13 & 269K & 219 & 55K & {\bf 103} & 129K & 116 & 174K & 149  & 55K & {\bf 105} \\

\hline





qwh-20-0 & 94K & 191 & 10K & {\bf 46} & 46K & 96 & 39K & 80 & 10K & {\bf 46} \\

\hline



qwh-20-2 & 869K & 1813 & 113K & {\bf 487} & 407K & 878 & 501K & 1025 & 113K & 490 \\

\hline

qwh-20-4 & 231K & 496 & 72K & 154 & 91K & {\bf 101} & 183K & 209 & 91K & 102 \\

\hline

qwh-20-5 & 89K & 176 & 32K & 118 & 41K & {\bf 70} & 67K & 124 & 41K & {\bf 69} \\

\hline


\end{tabular}
\end{footnotesize}
\end{center}
\vspace{-5mm} 
\label{table:all}
\end{table}

\begin{figure}[htb]
\begin{tabular}{cc}
\includegraphics[width=0.50\columnwidth]{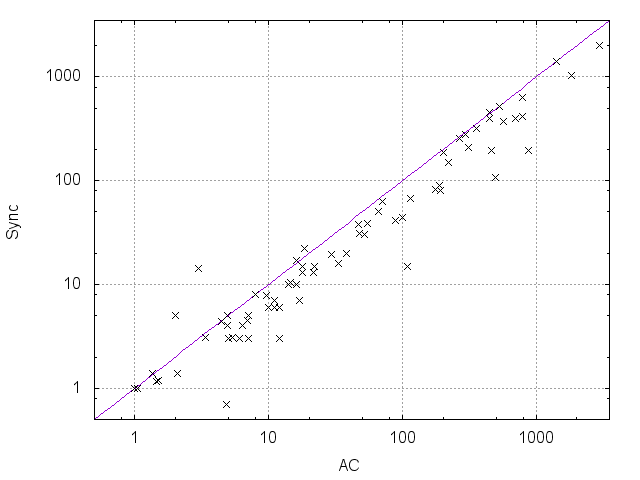} 
\includegraphics[width=0.50\columnwidth]{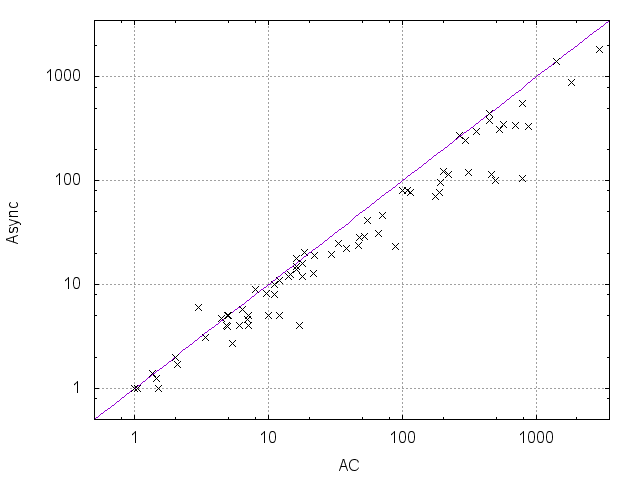}
\end{tabular}
\caption{Comparing Sync (left) and Async (right) to AC with respect to cpu time.} \label{syncACcpu}
\end{figure}

\begin{figure}[htb]
\begin{tabular}{c}
\includegraphics[width=0.50\columnwidth]{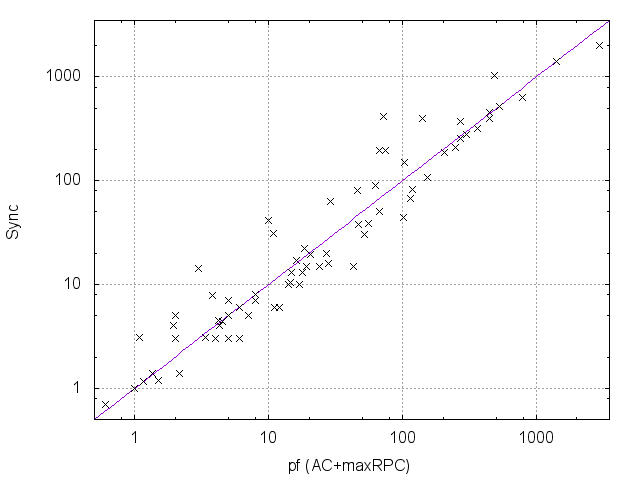} 
\includegraphics[width=0.50\columnwidth]{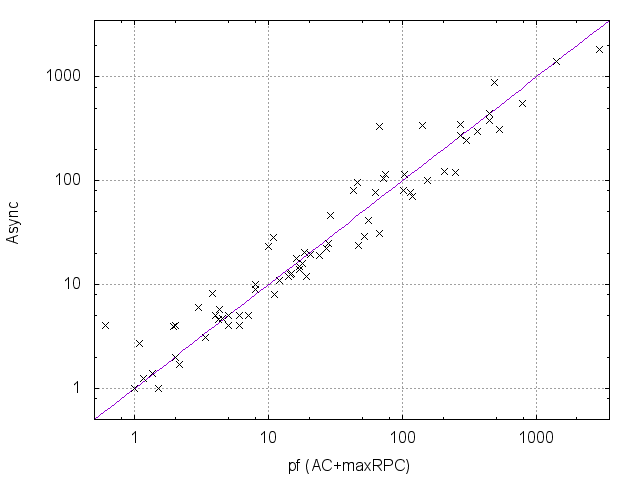} 
\end{tabular}
\caption{Comparing Sync (left) and Async (right) to {\em pf$_{AC+maxRPC}$} with respect to cpu time.} \label{syncportcpu}
\end{figure}

Table \ref{table:all} compares the performance of all the tested methods on various problem instances. Figures \ref{syncACcpu} and \ref{syncportcpu} give pairwise comparisons between our algorithms and AC (resp. pf$_{AC+maxRPC}$) by showing 
cpu times in log scale. We exclude very easy instances that are solvable by all methods in less than a second and very hard ones where the time limit was reached by all methods. Any point above (resp. below) the diagonal corresponds to an instance where AC/pf$_{AC+maxRPC}$ was better (resp. worse) than {\em Sync}/{\em Async}. 

The results demonstrate the validity of the motivation behind this work. 
As Figures \ref{syncACcpu} and \ref{syncportcpu} demonstrate, our algorithms are able to almost always outperform AC by taking advantage of the extra filtering offered by maxRPC without slowing down search, since this strong LC is applied in parallel to the main solver. Considering each problem class separately, we can note the following: 

\begin{itemize}

\item 
The rlfap is a class where AC dominates maxRPC as can been by the results of pf$_{AC+maxRPC}$ in Table \ref{table:all}, which usually follow those of AC. In this class 
both Async and Sync outperform AC, and therefore also pf$_{AC+maxRPC}$, with the latter algorithm being more efficient. Specifically, Sync can be up to twice as fast as AC and pf$_{AC+maxRPC}$.

\item
In gc problems AC is again better on average than maxRPC, but there are quite a few cases where the opposite occurs. In these problems {\em Async} and {\em Sync} are in most cases more efficient than AC and pf$_{AC+maxRPC}$, and there is no clear winner between them.

\item
AC is clearly better than maxRPC on driver and frb problems. The performance of {\em Sync} and {\em Async} is usually close to that of AC, but there are instances where they clearly outperform it.

\item
In the qcp and qwh classes maxRPC is typically by far faster than AC, very often by exponential margins. This is because of the considerable extra pruning it achieves. In most cases {\em Sync} and {\em Async} cannot match the performance of maxRPC and are therefore less efficient than pf$_{AC+maxRPC}$. However, the pruning achieved by the application of maxRPC inside {\em Sync} and {\em Async} is enough to make them clearly more efficient than AC. Also, there are some instances where our algorithms, and especially {\em Async}, are able to outperform maxRPC. 

\end{itemize}

Overall, we can say that the performance of {\em Sync} is perhaps surprising. Despite the limited time given to maxRPC while AC runs on the master process, it is able to achieve considerable extra pruning. This is reflected on 
cpu times where {\em Sync} is often twice as fast as AC, while it is rarely outperformed. 
Regarding the portfolio {\em pf$_{AC+maxRPC}$}, {\em Sync} is usually better on problems where the winner among the portfolio's solvers is AC (rlfap, gc, dr, frb). On problems where the winner is maxRPC (qcp, qwh) {\em pf$_{AC+maxRPC}$} is in most cases better than {\em Sync}.

Regarding {\em Async}, which has the greater potential for further development, results demonstrate that its performance follows that of maxRPC in terms of search space reduction, but without penalizing cpu times on problems where maxRPC is not successful. That is, while {\em Async} is faster than AC in most cases, it performs much better in problem classes where maxRPC excells. In qcp it can be up to 7 times faster than AC while  being always better than {\em Sync}. On the other hand, the differences between {\em Async} and AC are not very significant on rlfap, while on graph coloring AC is slightly better on some instances. 

These results are explained by looking at the backjumps that take place in each problem when solved by {\em Async}. The number of backjumps, as well as the mean and maximum numbers of search tree levels that are jumped over, are considerably higher in quasigroup instances compared to some other classes. For example, on qcp15-5 there were 7547 backjumps on average, and their mean and maximum lengths (i.e. $\#$levels jumped over) were 2.38 and 17.35 respectively. On driver-9, where {\em Async} is also very successful, there were only 52 backjumps on average but their mean and maximum lengths were 4.8 and 42.1. In contrast, on s11-f09 there were 2113 backjumps on average, and their mean and maximum lengths were 1.26 and 2.75. 

Finally, it is clear from Table \ref{table:all} that the portfolio {\em pf$_{all}$} which includes all four methods outperforms each individual method, as well as the portfolio {\em pf$_{AC+maxRPC}$}. For each given instance, {\em pf$_{all}$} matches the performance of the algorithm that performs best on this instance among AC, maxRPC, {\em Sync}, and {\em Async}. Therefore, it is the clear winner overall.
It is well known that when building portfolios using different solvers or a single solver under different parameter settings, a very important desired attribute is variability. That is, in order for the portfolio to be successful the solvers included in the portfolio should display dissimilar behavior. This is achieved by {\em pf$_{all}$} because, as demonstrated in Table \ref{table:all}, any method among {\em pf$_{AC+maxRPC}$}, {\em Sync}, and {\em Async} can be the winner on different instances and problem classes. Hence, they display variability in their performance, and this is why their integration into {\em pf$_{all}$} is very successful.

\section{Related Work}

\label{sec:related}

Obviously, the proposed framework for the efficient application of strong LCs 
requires the availability of machines with more than one processor (core). Since such machines are the norm nowadays, this is by no means a prohibitive requirement. In addition, the implementation effort required is very small, given already 
implemented algorithms for strong LCs.

There is a quite extensive body of work on parallel constraint solving which aims at exploiting the increasing number of available processors to speed up computation. A review can be found in the paper by \cite{Review2011}. Such works are relevant to our framework since they also exploit multiple processors, but at the same time they are quite different. 
Parallel CSP (and SAT) solving has mainly focused on search space splitting (i.e. allocating different branches of the search tree to different processors), e.g. the works by \cite{Perron99,Jaffar2004,MichelSH09,ChuSS09,Bordeaux2009,regin2013}, 
and solver portfolios, e.g. the works by \cite{HamadiJS09,HyvarinenJN09,AudemardS14,YunCP12,Dasygenis2014}, and to a lesser extent, on the parallelization of propagation, e.g. the works by \cite{Kasif90,Ruiz98,Rolf2010}. 

Regarding the latter direction, which is closer to our work, parallelizing constraint 
propagation algorithms is a challenging task since most such algorithms are sequential by nature, as demonstrated by \cite{Kasif90}. Hence, this approach has not been explored as much as the other ones, and it is quite different to our work where each LC algorithm runs on a single processor. Another common perception that has resulted in limited research on constraint propagation parallelization is that the scalability of this approach is limited by Amdahl's law: "if propagation consumes 80$\%$ of the runtime, then by parallelizing it, even with a massive number of processors, the speed-up that can be obtained will be under 5", as \cite{Bordeaux2009} explains. 

Existing works on parallel constraint propagation 
have focused on AC and have either been purely theoretical, or any experiments that were conducted, e.g. by \cite{Ruiz98} and \cite{NguyenD98} either 
failed to show significant speed-ups or were limited to very few processors. 
\cite{Rolf2010} consider the parallelization of a modern CP solvers' constraint propagation engine and shows that problems with a large number of (expensive to propagate) 
global constraints can benefit from parallelization of the propagation mechanism. 
Since this approach is orthogonal to ours, their combination is an interesting avenue for research. 

Our work is orthogonal to search space splitting methods since our algorithms explore a single search tree, on the master processor, and use a number of slaves to help speed up the exploration of this tree. However, it is feasible to combine our approach with search space splitting by first allocating different branches to different processors and then committing a number of slaves to each of these processors, along the lines of our framework.

Running a portfolio of solvers where each one applies a different LC is not the same as using these LCs within our framework, as our experimental results demonstrate. For example, consider a simple portfolio of two solvers where the first maintains AC and the second a stronger LC. It is quite likely that on some problem the first solver thrashes while the second explores a much smaller tree but spends too much time applying the strong LC at every node. In contrast, algorithm {\em Sync} may exploit the applications of the strong LC, even if its fixpoint is not reached, to remove some extra values and quickly direct search on the master process to a fruitful area of the search tree. Also, the application the strong LC within algorithm {\em Async} may result in large backjumps and thus avoid thrashing.

Finally, we need to note that through the use of strong local consistencies for propagation the scalability limitation posed by Amdahl's law can be overcome. This is because such consistencies we can achieve significantly stronger pruning than standard methods (such as AC), and therefore in many cases we can result in exponentially smaller search trees and corresponding run times.

\section{Conclusions}

\label{sec:conclusions}

We presented two novel ways to exploit the filtering power of strong LCs without paying a severe cpu time cost when they are not successful. Algorithm {\em Sync} 
applies strong LCs in parallel to the main propagation mechanism of the solver at each node of the search tree. The LC algorithms are stopped once propagation on the main process terminates, or if some LC algorithm detects a failure. 
Algorithm {\em Async} can apply different LCs at different nodes of the search tree at the same time. This can result in non-chronological backtracks of the main solver. 
Initial experimental results demonstrate the potential of our methods. We believe that the work presented here can open up numerous possibilities of parallelizing constraint propagation. Another important contribution of this study is to further motivate research on strong local consistencies and propagation methods. 


\end{document}